\pdfoutput=1

\documentclass[11pt]{article}

\usepackage[final]{acl}

\usepackage{times}
\usepackage{latexsym}
\usepackage{graphicx}
\usepackage{caption}
\usepackage{float}

\usepackage[T1]{fontenc}

\usepackage[utf8]{inputenc}

\usepackage{microtype}

\usepackage{inconsolata}

\usepackage{listings}
\usepackage{float}
\newfloat{lstfloat}{htbp}{lop}
\floatname{lstfloat}{Listing}

\title{Ace-CEFR - A Dataset for Automated Evaluation of the Linguistic Difficulty of Conversational Texts for LLM Applications}

\author{
  David Kogan \quad Max Schumacher \quad Sam Nguyen \\
  {\bf Masanori Suzuki \quad Melissa Smith \quad Chloe Sophia Bellows} \\
  {\bf Jared Bernstein} \AND
}

\begin{document}
\maketitle
\begin{abstract}

There  is  an  unmet  need  to  evaluate  the  language difficulty of short, conversational passages of text, particularly for training and filtering Large Language Models (LLMs). We introduce Ace-CEFR, a dataset of English conversational text passages expert-annotated with their corresponding level of text difficulty. We experiment with several models on Ace-CEFR, including Transformer-based models and LLMs. We show that models trained on Ace-CEFR can measure text difficulty more accurately than human experts and have latency appropriate to production environments. Finally, we release the Ace-CEFR dataset to the public for research and development.

\end{abstract}

\section{Introduction}

In the domain of language acquisition tools, a key capability is the measurement of the linguistic difficulty of text. Traditionally, this has been used to assess a language learner's ability by evaluating their writing \cite{arnold2018predicting,10.1007/978-3-030-29736-7_23,kerz-etal-2021-automated}. With the advent of use of Large Language Models (LLMs) for language learning and practice \cite{bonner2023large,kwon2023interfaces,mahajan-2022-bela,young2023investigating}, a novel application has arisen: adjusting the language output of an LLM to the ability of a specific user. This can be used to adjust content to a user's level of understanding, or to maximize a user's learning by keeping them in the Zone of Proximal Development (ZPD) \cite{kinginger2002defining}, reducing the difficulty for beginners and increasing it for more advanced users.

While LLMs have some innate understanding of text complexity, this typically takes the form of text simplification, especially on long text passages \cite{cardon2023operations,espinosa-zaragoza-etal-2023-review}. In contrast, language learning requires exposure to short, authentic text segments \cite{leow1997effects}, such as conversation. While LLMs are uniquely positioned to provide this, they are not typically trained to generate text at a learner's level.

\begin{figure}[H]
\centering
\includegraphics[width=7.7cm]{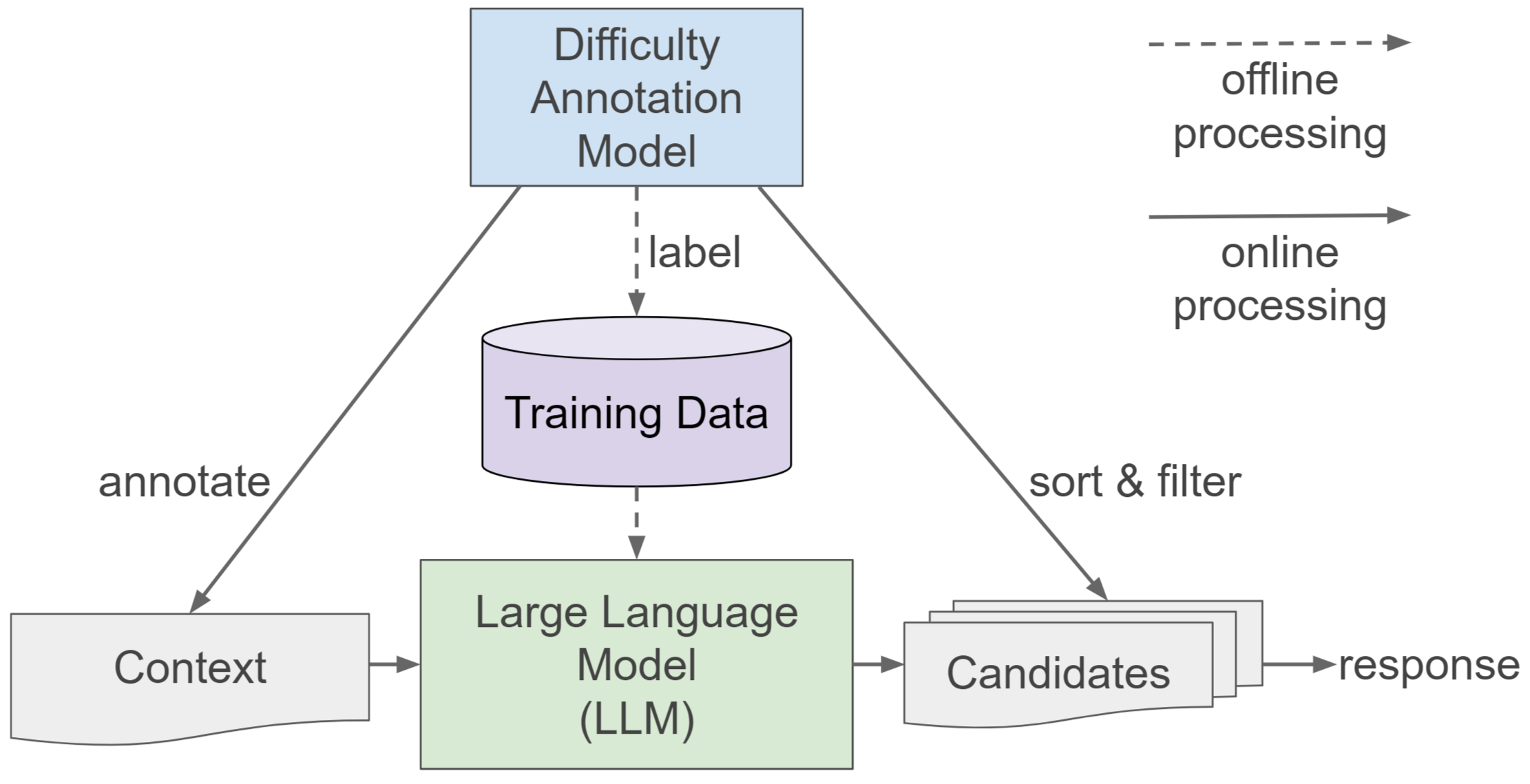}
\caption{Example system diagram of LLM trained to produce text at different levels of difficulty, with a Difficulty Annotation Model required to label text at three points in the processing pipeline.}
\label{fig:llm_overview}%
\end{figure}

To generate difficulty-tuned text directly, LLMs need offline and online modules that are able to evaluate such texts. In this kind of system, a difficulty model is used to label training data, annotate prompts, and filter output. An example system of this kind is shown in Figure~\ref{fig:llm_overview}. Such applications require a mix of offline and online processing, with the latter being highly sensitive to latency.

To be effective in this kind of system, the difficulty annotation model must be trained on texts analogous to those the LLM is generating, which means short, conversational passages.

\subsection{Summary of Contributions}

\begin{itemize}
    \item We release a novel dataset, Ace-CEFR, for English language difficulty. The dataset can be used to train models to understand the difficulty of text, as well as to train LLMs to generate text at specified levels, or for related tasks such as complex word identification.
    \item We establish baselines for performance on the difficulty evaluation task, for both human experts and machine models of different levels of complexity.
    \item We demonstrate the feasibility of medium size models to use the Ace-CEFR dataset to achieve good accuracy on the difficulty evaluation task, with latency suitable for real-time applications.
\end{itemize}

\subsection{Related Work}
\label{sec:related_work}

\subsubsection{Datasets}

There are a number of longer passage difficulty-annotated text datasets. These sets are comprised of passages on the order of hundreds of words in length each. These include the English First Cambridge open language Database (EFCAMDAT) \cite{Geertzen2014AutomaticLA}, the Cambridge Learner Corpus for the First Certificate in English (CLC-FCE) \cite{openclc}, Weebit \cite{rama2021pre}, OneStopEnglish \cite{vajjala2018onestopenglish}, Newsela \cite{nushi2020newsela}, a dataset provided by Adam Montgomerie \cite{amontgomerie}, Wiki-Auto \cite{jiang2020neural}, and the Sentence Corpus of Remedial English (SCoRE) \cite{chujo2015corpus}. These texts are deliberately long to establish a representative sample of difficulty \cite{shatz2020refining}.

The passages in these datasets are too long to train LLMs to produce conversational responses, being hundreds or more words long, compared to the average conversation turn length of approximately 10 words \cite{yuan2006towards}. We cannot simply split the passages up and train models on sub-passages, as individual sentences vary greatly from the overall passage assessment \cite{arase2022cefr}.

There are a few datasets annotated at the sentence level. These include \citet{vstajner2017automatic}, \citet{brunato2018sentence}, \cite{mcdonald2013universal}, and the CEFR-SP dataset \cite{arase2022cefr}.

However, these shorter datasets are not conversational, so are unsuitable for training a conversational LLM. As a representative example, the CEFR-SP dataset is composed of uniform, single-sentence, complete-thought sentences, and do not include the variations typically seen in conversations such as phrases, single word responses, references to other parts of the conversation, or multiple sentences.

Further difficulties in training models on all of the above datasets arise from unbalanced distributions of difficulties. The datasets are taken either from examples authored by language learners (e.g. EFCAMDAT and CLC-FCE), or sampled from natural text (e.g. CEFR-SP). This results in distributions that are highly skewed either toward the beginner or the middle of the difficulty curve, with almost no examples at high levels. This makes it difficult to train models capable of a wide range of evaluation. It is further worth noting that, while examples authored by language learners are ideal for evaluating learners, they are inappropriate for training LLMs to generate native-sounding speech.

For these reasons, we decided to author and annotate a novel dataset, composed deliberately of short, conversational texts at a variety of levels, including single words, phrases, sentences, and short passages.

\subsubsection{Modeling}

A variety of automated models have been used for the evaluation of text difficulty, typically using either readability scores or the Common European Framework of Reference \cite{cefrscale} scale, a standardized measure of language difficulty for L2 learners.

For readability, there are multiple defined metrics \cite{Matricciani2023ReadabilityID}, focused on the length and complexity of sentences and words. Readability prediction models measure those features, sometimes additionally considering word frequency statistics \cite{stenner1988lexile,fry1990readability,chall1995readability}, \citet{petersen2009machine} and word complexity \cite{aleksandrova-pouliot-2023-cefr} \cite{North_2023}. Recent works show that neural network-based approaches outperform statistical feature-based methods when using these features \cite{azpiazu2019multiattentive,meng2020readnet,imperial2021bert}, \cite{martinc2021supervised}.

However, readability is only representative of one kind of difficulty, and many research efforts focus on the CEFR scale, which evaluates multiple dimensions of difficulty, especially for L2 learners. \citet{salamoura2010exemplifying,ishii2018investigating} explored aligning English vocabulary and grammar with CEFR levels. \citet{uchida2018assigning} experimented with automated CEFR level assessment at the passage level, using data from Cambridge English exams. Notably, \citet{rama2021pre} showcased the high accuracy of Bidirectional Encoder Representations from Transformers (BERT) \cite{bert} in multilingual CEFR-level classification tasks, and \citet{arase2022cefr} developed a text CEFR level assessment model with BERT embeddings that performs significantly better than models based on superficial text features.

In alignment with these efforts, we have focused our modeling on the CEFR scale, applied to the Ace-CEFR dataset. To establish a clear baseline for further work, we evaluated a representative range of models, including statistical feature engineering, neural networks, and LLM prompting, analyzing their respective characteristics.

\section{Ace-CEFR Dataset}

To address the lack of short, conversation datasets described in Section \ref{sec:related_work}, we created a new dataset, targeting conversational texts, labeled by human language experts.

The Ace-CEFR (Annotated ConvErsational CEFR-aligned) dataset is comprised of 890 short text passages in English, created specifically for this task. The average length of a passage is 12 words, with a median of 10, aligned with typical conversation turn length \cite{yuan2006towards}. There are 62 passages composed of a single word each, and the longest passage is 114 words. While the dataset is small compared to some of the datasets in \ref{sec:related_work}, its properties make it possible to train models surpassing human experts (\ref{fig:mse}).

The dataset is comprised of a mix of sources: generated by our research organization for other language practice efforts (272), authored specifically for this task (255), generated by LLMs (198), pulled from conversations with trusted tester language learners, with anonymizations (101), and pulled from public data from the web (64). Anonymized conversation segments were processed via automated tools to remove potentially identifying information, and then further manually inspected and rewritten to ensure privacy. Much of the dataset is selected to be conversational in nature, since that is the primary expected application.

The texts were labeled aligned with the Common European Framework of Reference \cite{cefrscale} scale, a standard that organizes proficiency into six levels: A1-A2 (beginner), B1-B2 (intermediate), and C1-C2 (expert). In order to include examples of all levels, the dataset was labeled in batches of around 100, with a sampling method adjusted with the goal of a uniform distribution of levels. The distribution of floor(label) is A1:131, A2/A2+:180, B1/B1+:169, B2/B2+:186, C1:107, C2:116. Subsampling techniques can be used to achieve a perfectly balanced distribution if needed.

For the C1 and C2 levels, language experts created examples using both advanced vocabulary (e.g., ``He feigned indifference.'') and colloquial and idiomatic usage (e.g., ``Get off your high horse and lend me a hand. This house isn't going to paint itself.'')

\subsection{Human Expert Labels}
\label{headings}

Passages in the dataset were rated by English language learning experts, each with at least a Master's degree in Applied Linguistics or similar, plus a minimum of 10 years of experience in language teaching, language teaching curricula and assessment development, teacher education, or research in the field. Labels were applied on the CEFR scale \cite{cefrscale}: A1 through C2. By convention, the labels A2 through B2 include ``+'' variations, indicating a level higher than the baseline. 

Each text was labeled by at least two raters, working independently, but collaborating on a rating guideline document to align themselves. The CEFR labels were applied based on the productive difficulty, i.e., the level at which an L2 learner can be  expected to produce the text. For texts composed of a single homograph, the meaning with the lowest level was chosen, as that is most likely to be used by a language learner.

Ratings were converted to numbers (A1=1, A2=2, A2+=2.5, B1=3, B1+=3.5, B2=4, B2+=4.5, C1=5, C2=6), and averaged to arrive at a consensus per text. In some cases, more raters were available and we included those in the average (112 cases). 

While most human expert labels were within 1 point of one another, 8\% of the labels were further apart than this. Disagreements were particularly common for intermediate CEFR levels, but the quadratic weighted kappa (QWK) between the two primary raters is 0.89, which indicates overall close agreement.

In about 5\% of cases, due to differences greater than 1 between individual raters, labels were adjudicated by expert raters as a group to arrive at a consensus label. At the end of model training for each of the Linear, BERT-based and PaLM 2-L models, the worst 20 predictions from each were re-adjudicated to identify potential mislabels. Results presented in the Experiment section (section \ref{sec:experiment}) are on the final dataset, after all adjudication was completed (123 cases of adjudication in total).

\section{Evaluation Framework}
\label{headings}

We split the ACE-CEFR evenly into training (445) and test (445) sets. The same train and test set was used for all models.

We evaluated our the on predicting the labels in the test set. Because of averaging between raters, the labels are not constrained to CEFR boundaries, e.g., ``I have lived here since I was 4.'' is labeled 2.75, meaning that it falls between the A2+ and B1 CEFR labels. Our primary metric was therefore chosen to be Mean Squared Error (MSE) between a model's predictions and the consensus human expert label, on the 1-6 scale, meaning the maximum error possible is 5, and accordingly the maximum MSE is 25.

In addition to accuracy, latency is a major practical consideration. Some use cases, like generating offline training data, are relatively latency insensitive, but others are in the critical path, like integrating with an LLM for generation (Figure~\ref{fig:llm_overview}) or evaluating user proficiency in real time. For key applications, a model with latency in the 10ms to 100ms range is desirable.

\section{Experiment}
\label{sec:experiment}

\subsection{Models Overview}

We evaluated three types of models, in order from simplest to most complex: a linear regression model on surface language features, a custom model fine-tuned off Bidirectional Encoder Representations from Transformers (BERT) \cite{bert}, and a Large Language Model (PaLM 2-L) \cite{anil2023palm} in a few-shot setting. Fine-tuning an LLM was not a focus of this research due to its higher cost and limited accessibility to most LLM users \cite{trad2024prompt} \cite{Xu2023WizardLMEL}, but it is a topic of interest for future investigation. As a comparison baseline, the test set was also rated by a human expert. 

\begin{figure*}[ht]
\centering
\includegraphics[width=16cm]{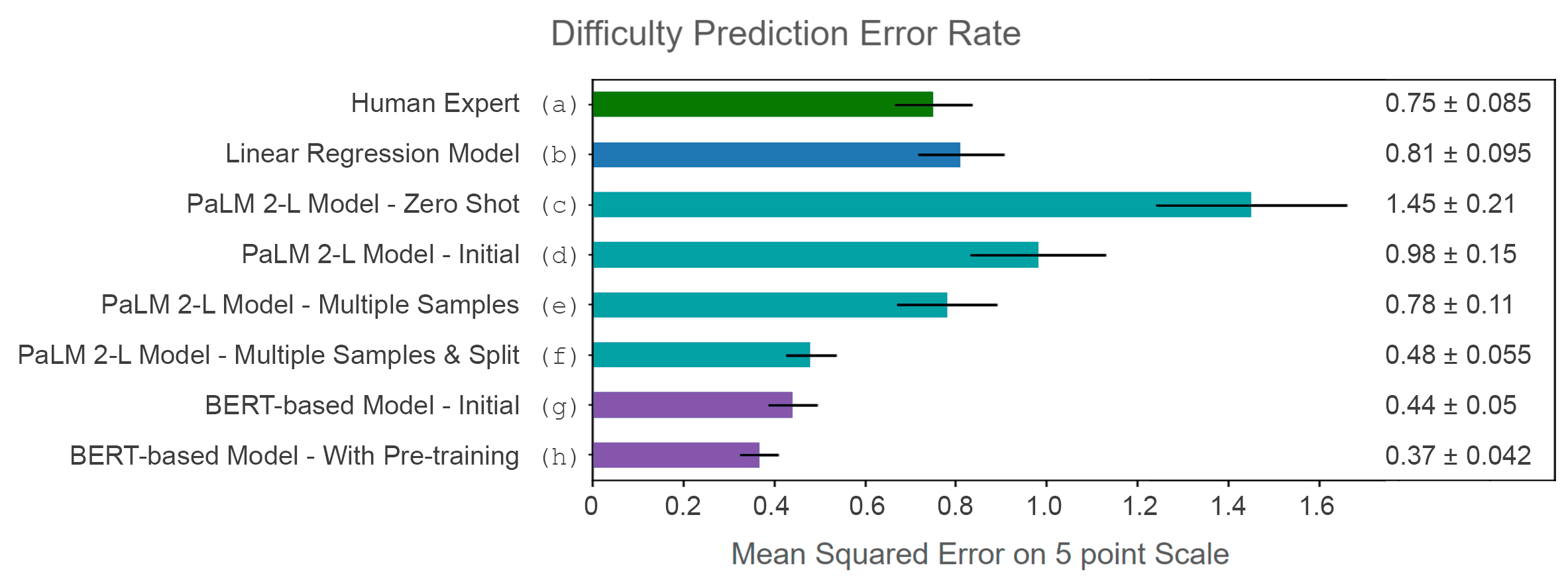}
\caption{Summary of mean squared error using the Ace-CEFR set to train difficulty prediction models, with 90\% confidence intervals. See Section \ref{sec:experiment} for detailed results and analysis.}
\label{fig:mse}%
\end{figure*}

Summary of MSE results is in Figure~\ref{fig:mse}, and latency results are in Table~\ref{tab:latency}.

{\renewcommand{\arraystretch}{1.5}%
\begin{table}[h!]
  \scriptsize
  \begin{center}
    \caption{Latency summary of single lookup latency averaged over 100 requests. Latency is estimated within an order of magnitude, and no effort has been made to optimize code for speed. CPU latency was measured on a Linux desktop Intel(R) Xeon(R) CPU E5-2690 v4 @ 2.60GHz with 128 Gb RAM. TPU latency was measured via the Vertex API on a low-latency network connection, querying TPU v5e accelerators. Note that TPU execution is highly parallelizable, so amortized batch lookup speed is substantially faster than individual lookup.}
    \label{tab:latency}
    \begin{tabular}{p{1in}|c|c}
     \textbf{Model Type} & \textbf{Method} & \textbf{Latency}\\
      &  & (One lookup)\\\hline
     Linear Model on Surface Features & On-device (CPU) & $\sim$50$\mu$s \\\hline
     BERT-based Model & On-device (CPU) & $\sim$100ms\\\hline
     BERT-based Model & Via API & $\sim$10ms\\\hline
     PaLM 2-L & Via API & $\sim$1s\\\hline
    \end{tabular}
  \end{center}
\end{table}
}

\subsection{Human Expert}
\label{sec:human_expert}

As a basis for comparison, a set of ratings was performed on the test set by a human expert with the same qualifications as the original raters. This expert did not previously work with the labelers of the dataset, but used the rating guideline as well as the training set labels for calibration. Their labels had a MSE of 0.75 (90\% confidence [0.67, 0.84]) (Figure~\ref{fig:mse} (a)). 

\subsection{Linear Regression Model}
\label{headings}

The benefit of such models is their simplicity and speed. The model we built can execute locally in-process, with latency measured in microseconds. The downside is that their accuracy is limited because of a lack of text understanding.

\subsubsection{Features}
\label{headings}

There is considerable prior research on measuring text difficulty using surface features such as sentence and word length \cite{khushik2022syntactic} and word diversity \cite{treffers2018back}. While these are not encompassing metrics of text complexity \cite{tanprasert2021flesch}, they correlate strongly with difficulty. After experimentation, we settled on the signals ``average word length in characters,'' ``average sentence length in characters,'' and ``average sentence length in words'', with correlations of 0.67, 0.70 and 0.35 to the difficulty. The sentence length signals has a logarithmic relationship to the difficulty, and taking ln(signal) improves the correlation to 0.71 for length in words and 0.75 for length in chars.

The key weakness of these features is that they are content agnostic. For example, ``The cat is here.'' (A1 difficulty) and ``His ire is epic.'' (C1/C2 difficulty) have indistinguishable word and sentence features. For these reasons, such approaches are most effective when averaged over long texts, and are much less useful for short conversational passages.

\subsubsection{Results}

The linear model got an MSE of 0.81 (90\% confidence [0.71-0.91]) (Figure~\ref{fig:mse} (b)). This is slightly worse than human expert labels, but within the confidence interval. Typical errors relate to mistaking the difficulty of a short word and sentences comprised of short words (Table~\ref{tab:linear_errors}). It also tends to overestimate the difficulty of sentences that are simple in structure, but have many words, e.g., ``For herbal tea, we have blueberry chamomile, chai, rooibos, fennel tarragon, and nettle.'' is labeled at 3 (B1) but predicted by the model to be 5 (C1).

\subsection{Large Language Model}

An LLM is a natural choice for evaluating the difficulty of text. Such models have intrinsic understanding of language, and their training data often organically includes the CEFR scale \cite{yancey2023rating}. It is possible to ask an LLM to evaluate text and get a reasonable response. The downside is that these models are comparatively slow (Table~ \ref{tab:latency}) and are therefore primarily suitable for offline text labeling.

We used the PaLM 2-L model \cite{anil2023palm}, a model optimized for language understanding, generation, and translation tasks. We limited ourselves to few-shot prompt engineering. It is likely that prompt tuning or fine tuning would yield better results, and this is a direction for future research.

\subsubsection{Results}

As a baseline, we first tested a zero-shot version, where we asked the model for a response without giving it any examples. This establishes the LLM's innate understanding of CEFR levels. This resulted in an MSE of 1.45 (Figure~\ref{fig:mse} (c)). While this is the worst of the results, it is notably better than a random guess (MSE of approximately 4.6) or always guessing the median of the training set (MSE of 2.37). This shows that the LLM does indeed have some understanding of CEFR levels, though an extremely imprecise one. Iterative improvements over this demonstrate the effectiveness of the Ace-CEFR training set.

For the initial few-shot results, we used a single prompt (\ref{sec:prompts}), populated by instructions and examples from the training data. Notably, because of the constraints of context length, we randomly sampled 64 out of 445 training examples. This resulted in an MSE of 0.98 (Figure~\ref{fig:mse} (d)).

Since the limitation of context length prevented us from using all of the training data as few-shot examples, we experimented with running the model multiple times, re-sampling the training data for few-shot examples, and averaging the results. By rerunning the model 3 times, we improved accuracy, from an MSE of 0.98 to 0.78 (Figure~\ref{fig:mse} (e)). Naturally, this results in proportionately increased latency. Further improvement is likely possible if more samples are taken. 

We noted that the model had significant difficulty predicting the label of single words compared to phrases. We hypothesized that this is because from the LLM's perspective, these are very different tasks, and because many more of the training examples are phrases (N=418) compared to single words (N=27). Since the training examples are further subsampled in sets of 64 to fit in the context, only 3-4 single words would actually be seen by the model.

To address this, we separated the prompts into two types: one responsible for predicting the difficulty of phrases, and another one for predicting the difficulty of individual words (Appendix~\ref{sec:prompts}).  This significantly improved the MSE, from 0.78 to 0.48 (Figure~\ref{fig:mse} (f)).

The final results are an MSE of 0.48 (90\% confidence [0.43, 0.54]) (Figure~\ref{fig:mse} (f)). This is 0.33 better than the linear model and 0.27 better than human expert ratings, albeit at a significant latency cost (Table~\ref{tab:latency}). Unlike the linear model, there is no obvious pattern of errors (Table~\ref{tab:ulm_errors}). The opacity of mistakes is a risk factor, since this can make it challenging to improve the model further.

\subsection{BERT-based Model}

The BERT-based model builds on an existing, lightweight BERT encoder, which provides a combination of a high degree of accuracy and production-level latency. We fine-tuned a custom model by taking the first few layers of the pretrained BERT-base-uncased checkpoint and adding a classification head. The BERT encoder is multiple orders of magnitude smaller than a typical LLM (millions rather than billions of parameters), but still comes pretrained with a degree of language understanding and is easily fine-tuned to very specific tasks. It is also well-suited to learn from a larger teacher model, which was used during a quality iteration.

\subsubsection{Results}

We finetuned the BERT encoder on the 445 training samples, and ran light hyperparameter tuning (on a validation set split from the training samples) for the number of layers of the pretrained encoder to keep learning rate and batch size. The best setup retained the first 3 layers, trained with a learning rate of $6e-5$ at batch size 32 for 6 epochs. The final model has 45.7M parameters and achieved an MSE of about 0.44 (Figure~\ref{fig:mse} (g)), which is substantially better than any of the other models. 

Unlike the linear model, which peaks in accuracy after a few dozen examples, and the LLM, which is context-constrained to accept only a few dozen examples, the BERT model continues to improve with additional training data. We therefore added an extra finetuning stage to the training. In the first stage, we labeled 10,000 examples from various sources with our best LLM version. We used those LLM-labeled examples to finetune the BERT model using a smaller learning rate of $2e-5$. In the second stage, we further finetuned the model on the human expert rated dataset. The results improved significantly, from MSE 0.44 to 0.37 (Figure~\ref{fig:mse} (h)).

The final results are an MSE of 0.37 (90\% confidence [0.32, 0.41)] (Figure~\ref{fig:mse} (h)), which is a 0.38 better than the human expert. The latency, particularly when running on TPU (Table~\ref{tab:latency}), is also practical enough for latency-sensitive production applications, making this the ideal model for most use cases.

The only recurring issue we saw was that this model struggled with misspellings, compared to the LLM (with its larger vocabulary) and the Linear Model (which has no concept of spelling). We did not deliberately introduce misspellings into the Ace-CEFR dataset, but they arose naturally from several of our sources. Ultimately, we decided to correct the misspellings, because we want the dataset to be usable for generative tuning, and mistakes in the input could cause an LLM to learn to produce misspellings. However, this is a weakness that needs to be taken into account when integrating into production use cases, and a spell-checker may be helpful.

Aside from misspellings, the BERT-based model's errors were similarly opaque to the LLM errors. The only significant pattern was having difficulty with idiomatic sayings, like ``It's been a rough spell but I'm game to try anything that might help us weather this storm.'' (Table~\ref{tab:bert_errors})

\subsection{Ensemble Models}

It is noteworthy that while each model makes mistakes, the categories of mistakes made by different models differ. For example, the Linear Model has no concept of semantics, whereas the BERT model has no concept of word length. We therefore evaluated whether it's possible to offset the errors of the different models by combining them together.

To do so, we randomly split out 100 examples from the test set to use for tuning, and used the remaining 355 examples for evaluation. We weighted the models to optimize performance on the tuning set, essentially putting a linear model over them. With this approach, we were able to reduce MSE from 0.36 for BERT to 0.33 when combining BERT+LLM. Adding the linear model to the mix did not improve results further beyond noise levels.

While this improvement is incremental, and likely incurs too much complexity to be used in production, it establishes that further improvements in accuracy are possible, and this approach may be useful for creating better pre-training datasets for improvements to BERT in the future.

\section{Conclusion}
\label{headings}

Ultimately, we were able to achieve accuracy better than expert human ratings on short conversational pieces of text. We are releasing the Ace-CEFR dataset to the public for further iteration, and believe it can be of significant use in building LLM-based language learning applications.

\section{Limitations}

The Ace-CEFR dataset provides the ability to train models on conversational text, but it has several  limitations. 

It was generated from a limited set of sources and rated by a small cohort of expert raters. Diversifying both the sources and the raters may provide significantly less biased and more generalized results. Additionally, the dataset, and all the models trained on it here, are limited to English, which does not serve populations trying to learn other languages. Expanding the dataset to other languages is possible, but would require incremental work per language unless an automated methodology is identified. Iterating on the dataset, either to expand it to other languages or to add more data to English, also requires considerable human expert work. While the co-authors of this paper had the expertise to do this directly, these are not always readily available skills. In many cases, vendor labor is used for this kind of work, which raises ethical concerns about fair compensation for such labor, and appropriate usage of results.

Another significant limitation of these approaches is that they rely on a single scale for difficulty, which is not representative of the diverse experiences and backgrounds of learners. Particularly impactful is the L1 of the learner, which greatly affects both overall learning difficulty and specific skill acquisition {\cite{Ellis1985UnderstandingSL}}. For example, because French and English have many more cognates than Arabic and English, an L1 speaker of French will likely find different areas of challenge when learning English than an L1 speaker of Arabic. This makes a single scale of difficulty for the two learners to be imperfect for either learner. A more fine-grained and personalized approach to user challenge is likely to be made possible by the advent of LLMs, and is a fertile ground for future research.
 
A broader inequity inherent to automated tools is the unequal availability of technology to learners of different demographics. Access to computers or mobile phones is not available to everyone, and the demographics that have the most difficulty getting traditional second language education are also likely the ones who will have the least access to computers and mobile phones capable of accessing LLM-based applications for learning. It is important to consider how to maximize accessibility when building applications on top of these technologies, for example, by making them compatible with entry-level consumer devices.
 
\section{Future Work}
\label{headings}

The next natural step is integrating this work into LLM generation, using both the manually-labeled difficulty dataset and the automated difficulty measuring models.

Additionally, there is considerable work to be done to improve the dataset, as mentioned in the Limitations section, including size, diversity, and scaling to non-English languages. 

Beyond that, there's still headroom to further improve accuracy, as demonstrated by the ensemble model experimentation. We believe that adding a dictionary of average word frequency or difficulty to the Linear model, such as the Global Scale of English dictionary \cite{GSE}, would significantly improve its results without sacrificing latency, though it's not expected it would surpass the neural-network based models. Such a dictionary could also be automatically generated using the larger models. Finetuning LLMs can also be insightful to compare the results against few-shot prompting. Other improvements could be using an LLM with a longer context to include more examples, and cross-training with other datasets such as CEFR-SP. Further work in distillation is also of great practical interest, particularly distilling LLM and BERT-based models into smaller versions with lower latency and operational costs.

\bibliography{custom}

\clearpage

\appendix

\section{LLM Prompts}
\label{sec:prompts}
\begin{lstfloat}[H]
\begin{lstlisting}[caption={Prompt to Evaluate Text Difficulty for Phrases (also initially used for words)},label={lst:prompt_phrase}]
CEFR is a six-level scale, with each level corresponding to a specific level of English language proficiency. The levels are: 

- A1 (1): Beginner
- A2 (2): Elementary
- B1 (3): Intermediate
- B2 (4): Upper Intermediate
- C1 (5): Advanced
- C2 (6): Proficiency

According to the CEFR scale, the proficiency level required to use the following phrases are:

Phrase: You are welcome! -> CEFR: 1
Phrase: I wonder if there's any treasure. -> CEFR: 3.25
[more examples...]
Phrase: {test_phrase} -> CEFR:
\end{lstlisting}
\end{lstfloat}

\begin{lstfloat}[H]
\begin{lstlisting}[caption={Prompt to Evaluate Text Difficulty for Single Words},label={lst:prompt_word}] 
GSE is a six-level scale, with each level corresponding to a specific level of English language proficiency. The levels are:

- A1 (1): Beginner
- A2 (2): Elementary
- B1 (3): Intermediate
- B2 (4): Upper Intermediate
- C1 (5): Advanced
- C2 (6): Proficiency

According to the GSE scale, the proficiency level required to use the following words are:

age,1
almost,2
[more examples...]
{test_word},
\end{lstlisting}
\end{lstfloat}

\clearpage
\section{Example Errors}
\label{sec:errors}

Tables with the worst error examples from each model type.

{\renewcommand{\arraystretch}{1.5}%
\begin{table}[h!]
  \scriptsize
  \begin{center}
    \caption{\textbf{Human Expert Rater}: worst 5 errors, labels are 1-6 with 1 corresponding to A1 on the CEFR scale and 6 corresponding to C2}
    \label{tab:linear_errors}
    \begin{tabular}{p{1.3in}|c|c|c}
      \textbf{Text} & \textbf{Label} & \textbf{Prediction} & \textbf{Error}\\\hline
      The Sumida River is one of Japan's biggest, and you can take a tour on a boat and see the sights along the river's edges like sumida aquarium, temples, and more. The Sumida Observatory lets you take in a birdseye view of the river and Tokyo. Are you ready to book your tickets? & 5 & 2.5 & -2.5\\\hline
      I have a nice garden with flowers, trees, and a small pond. & 3.25 & 1 & -2.25\\\hline
      I like the classics over remakes. & 4.75 & 2.5 & -2.25\\\hline
      I see. Dulce de leche is a popular dessert in Argentina, and it is often used as a filling for pastries and other desserts. Empanadas are also a popular dish in Argentina, and they can be filled with a variety of ingredients, such as meat, cheese, or vegetables. & 5.25 & 3 & -2.25\\\hline
      I'm looking to the future with hope. & 4.25 & 2 & -2.25\\
    \end{tabular}
  \end{center}
\end{table}
}

{\renewcommand{\arraystretch}{1.5}%
\begin{table}[h!]
  \scriptsize
  \begin{center}
    \caption{\textbf{Linear Model}: worst 5 errors, labels are 1-6 with 1 corresponding to A1 on the CEFR scale and 6 corresponding to C2}
    \label{tab:linear_errors}
    \begin{tabular}{p{1.3in}|c|c|c}
      \textbf{Text} & \textbf{Label} & \textbf{Prediction} & \textbf{Error}\\\hline
      to ascertain & 6 & 2.4 & -3.6\\\hline
      naive & 4 & 1.1 & -2.9\\\hline
      endeavor & 5 & 2.4 & -2.6\\\hline
      Get off your high horse and lend me a hand. This house isn't going to paint itself. & 6 & 3.6 & -2.4\\\hline
      effervescent & 6 & 3.6 & -2.4\\
    \end{tabular}
  \end{center}
\end{table}
}

{\renewcommand{\arraystretch}{1.5}%
\begin{table}[h!]
  \scriptsize
  \begin{center}
    \caption{\textbf{PaLM 2-L}: worst 5 errors, labels are 1-6 with 1 corresponding to A1 on the CEFR scale and 6 corresponding to C2}
    \label{tab:ulm_errors}
    \begin{tabular}{p{1.3in}|c|c|c}
      \textbf{Text} & \textbf{Label} & \textbf{Prediction} & \textbf{Error}\\\hline
      By perseverance. & 4 & 1 & -3\\\hline
      Just a couple of weeks. & 1 & 3 & 2\\\hline
      By perseverance, just not giving up even when things seem impossible. & 5.5 & 3.87 & -1.63\\\hline
      The rate at which kids absorb new information is simply astonishing. & 6 & 4.4 & -1.6\\\hline
      Yeah, it's quite a controversy! & 4.75 & 3.2 & -1.55\\
    \end{tabular}
  \end{center}
\end{table}
}

{\renewcommand{\arraystretch}{1.5}%
\begin{table}[h!]
  \scriptsize
  \begin{center}
    \caption{\textbf{BERT-based model}: worst 5 errors, labels are 1-6 with 1 corresponding to A1 on the CEFR scale and 6 corresponding to C2}
    \label{tab:bert_errors}
    \begin{tabular}{p{1.3in}|c|c|c}
      \textbf{Text} & \textbf{Label} & \textbf{Prediction} & \textbf{Error}\\\hline
      hobby & 1 & 3.23 & 2.23\\\hline
      Celery is a low calorie vegetable. & 4 & 2.13 & -1.87\\\hline
      I didn't understand the noise last night. & 2.25 & 3.82 & 1.57\\\hline
      I am definitely leaning towards accepting it. & 3.5 & 5.02 & 1.52\\\hline
      Get off your high horse and lend me a hand. This house isn't going to paint itself. & 6.0 & 4.55 & -1.45\\
    \end{tabular}
  \end{center}
\end{table}
}

\end{document}